\documentclass[journal]{IEEEtran}

\usepackage[noadjust]{cite}
\usepackage{graphicx}
\usepackage{amsmath}
\usepackage{amssymb}
\usepackage{multirow}
\usepackage[letterpaper=true,colorlinks,bookmarks=false]{hyperref}
\usepackage{subfigure}
\usepackage[lined,boxed, ruled]{algorithm2e}
\title{ Sparse Range-constrained Learning and Its Application for Medical Image Grading}

\author{Jun Cheng 
 \thanks{This work was supported in part by  Cixi Institute of Biomedical Engineering, Chinese Academy of Sciences, China, under Grant Y80002RA01. (Corresponding Author： Jun Cheng) }
\thanks{  J. Cheng is with Cixi Institute of Biomedical Engineering, Chinese Academy of Sciences, China (email: chengjun@nimte.ac.cn).}
 }
\begin{document}
 \markboth{ IEEE TRANSACTIONS ON MEDICAL IMAGING, vol. x, pp. x,
 ~2018}
  {Cheng \MakeLowercase{\textit{et al.}}:}

\maketitle

 \begin{abstract}Sparse learning has been shown to be effective in
 	solving many real-world problems. Finding sparse representations
 	is a fundamentally important topic in many fields of
 	science including signal processing, computer vision, genome
 	study and medical imaging. One important issue in applying
 	sparse representation is to find the basis to represent the data,
 	especially in computer vision and medical imaging where the
 	data is not necessary incoherent. In medical imaging, clinicians
 	often grade the severity or measure the risk score of a disease
 	based on images. This process is referred to as medical image
 	grading. Manual grading of the disease severity or risk score
 	is often used. However, it is tedious, subjective and expensive.
 	Sparse learning has been used for automatic grading of medical
 	images for different diseases. In the grading, we usually begin
 	with one step to find a sparse representation of the testing image
 	using a set of reference images or atoms from the dictionary. Then
 	in the second step, the selected atoms are used as references to
 	compute the grades of the testing images. Since the two steps
 	are conducted sequentially, the objective function in the first
 	step is not necessarily optimized for the second step. In this
 	paper, we propose a novel sparse range-constrained learning
 	(SRCL) algorithm for medical image grading. Different from
 	most of existing sparse learning algorithms, SRCL integrates
 	the objective of finding a sparse representation and that of
 	grading the image into one function. It aims to find a sparse
 	representation of the testing image based on atoms that are
 	most similar in both the data or feature representation and the
 	medical grading scores. We apply the new proposed SRCL to
 	two different applications, namely, cup-to-disc ratio computation
 	from retinal fundus images and cataract grading from slit-lamp
 	lens images. Experimental results show that the proposed method
 	is able to improve the accuracy in cup-to-disc ratio computation
 	and cataract grading.

 \end{abstract}
\section{Introduction}\label{introduction}
  Sparse learning is a representation learning method which aims at finding a sparse representation of the input data in the form of a linear combination of basic elements as well as those basic elements themselves. These elements are called atoms and they form a dictionary. Atoms in the dictionary are not necessarily  orthogonal, and they may be in an over-complete spanning set. 
Sparse  learning has been  shown to be effective in solving many
real-world problems.   Finding  sparse
representations  is a fundamentally important topic in many fields of
science including  signal
processing, computer vision, genome study and medical imaging. 
  In audio processing, sparse representation was used to decompose  sounds in terms of a dictionary of Gammatone functions   \cite{Adiloglu12}. In \cite{Yuanqing2004}, it was also used   for blind source separation and EEG signal analysis.  In image processing, sparsity has been shown to be promising in denoising and enhancement  \cite{BM4D, Fang13, Fang12, Elad06, Kafieh15, cheng14, chengjun16, feikou2013}.  In computer vision tasks such as face recognition \cite{Wright09, zhanglei2010,  LU2013111, Xing2013}, sparsity  has been widely used as well.  In genome-wide association studies, only a small portion of genes has been found to contribute to different diseases \cite{Golub99, zhangzhuo12, Li2015}. 
  In medical imaging, sparse learning based approaches have been used for brain  image analysis \cite{Aljabar2009MultiatlasBS, varoquaux:inria-00588898, Roy2015}. Recently, sparsity has been used  for medical image grading \cite{Xu13, CJ15}. 
  \begin{figure}
 	\centering
 	{\subfigure[Grade 1]{
 			\includegraphics[width=1.5in]{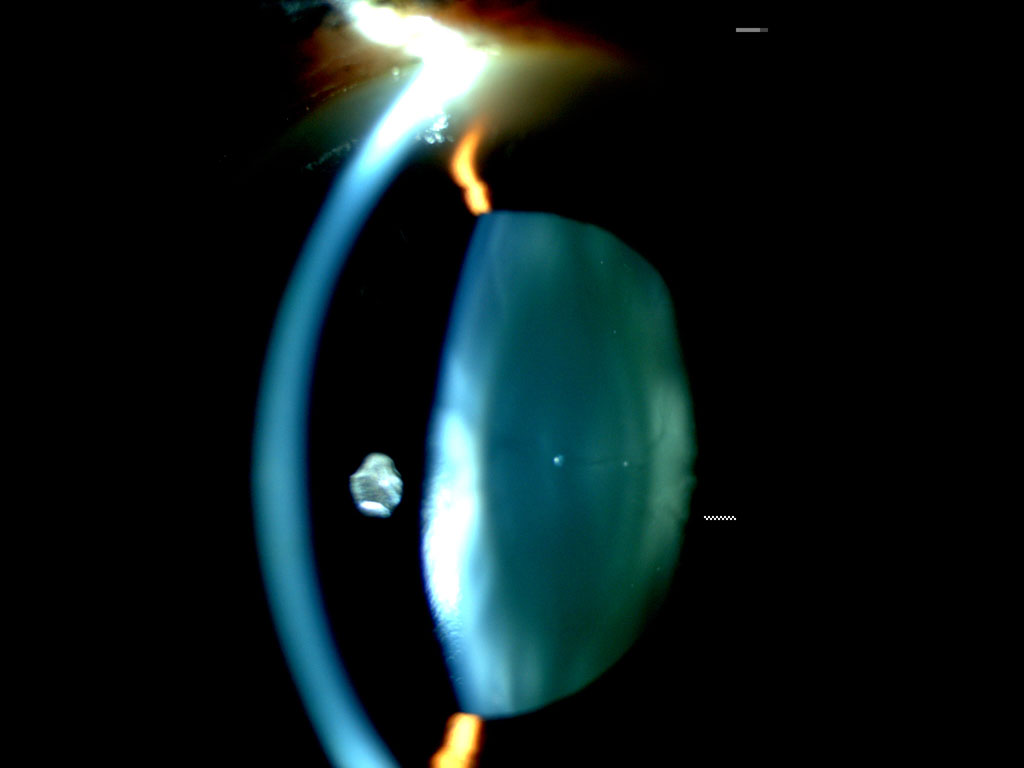}
 	}} 
 	{\subfigure[Grade 2]{
 			\includegraphics[width=1.5in]{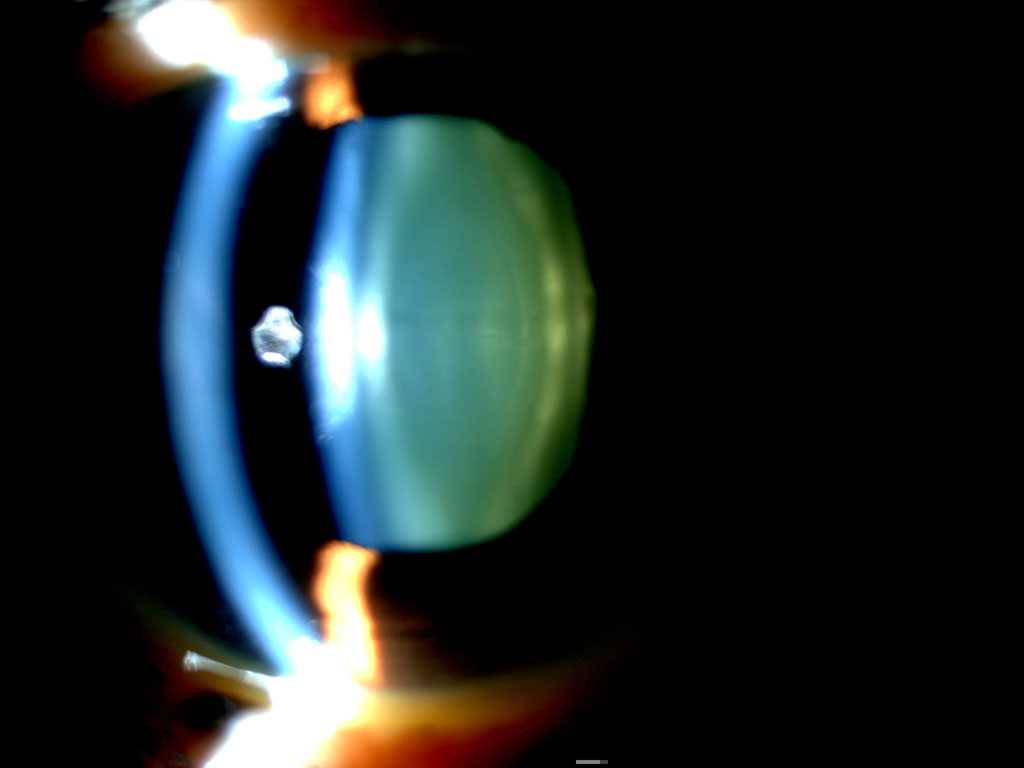}
 	}} 
 	{\subfigure[Grade 3]{
 			\includegraphics[width=1.5in]{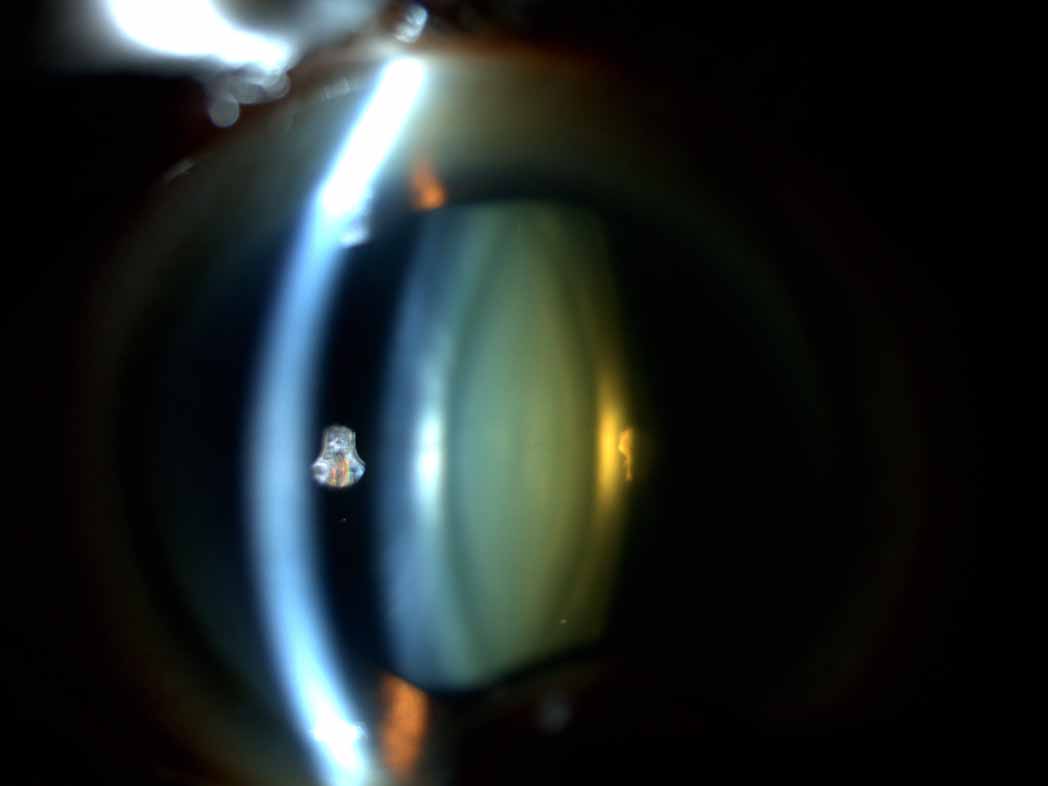}
 	}} 
 	{\subfigure[Grade 4]{
 			\includegraphics[width=1.5in]{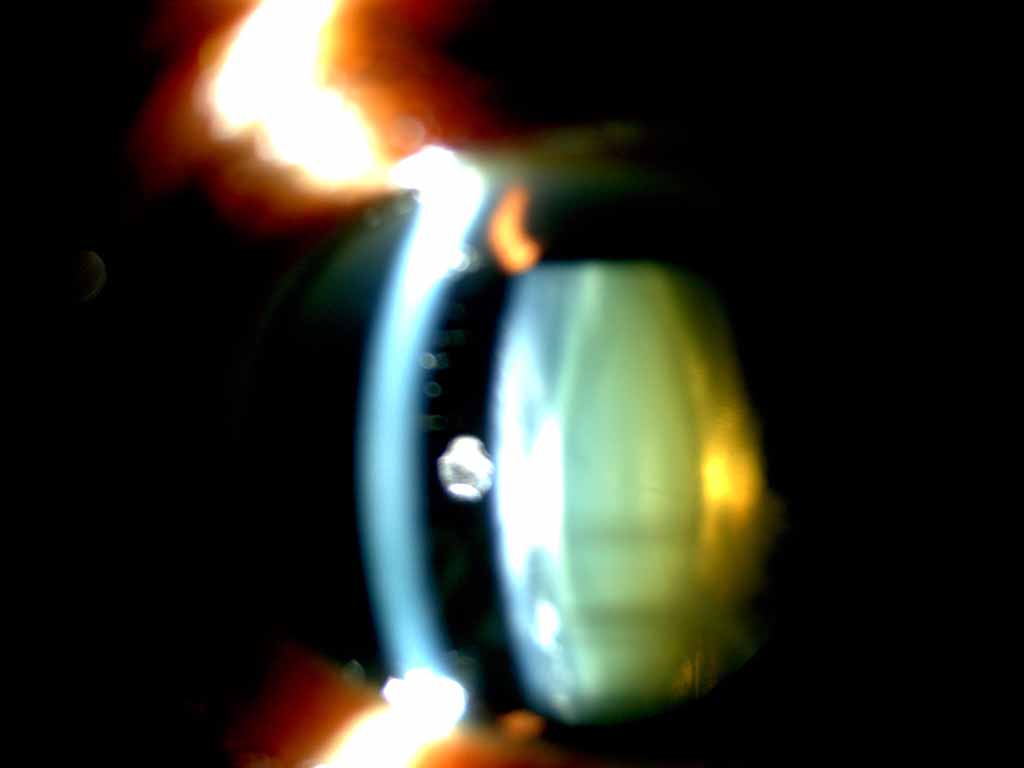}
 	}} 
 	\caption{Four slit-lamp images with manual grading from 1 to 4.} \label{fig0}
 \end{figure}
 
  Clinicians  often   grade the severity or measure the risk score of a disease  based on images. This process is referred to as medical image grading.  Very often, the image to be graded is compared with some reference images  based on protocols. For example, four standard images with grade 1 to 4, similar to that in Fig. \ref{fig0}, are provided as reference images in cataract grading \cite{Xu13}. Clinicians often compare a new image with  the standard images to get a decimal grading score. In diabetic retinopathy grading, the clinicians  examine the retinal images to determine the presence and types of  lesions, which would be used to determine the  severity or stage of the diseases. In glaucoma diagnosis, the clinicians measure the vertical cup-to-disc ratio (CDR) as a risk factor for glaucoma.   However, manual assessment or grading of the images is subjective, time-consuming and expensive. Therefore, automatic grading of the images is expected to improve the clinical management of the  diseases and provide an objective assessment for studies.

  Inspired by the manual grading process by clinicians, sparse learning has been proposed for medical image grading. 
  Let $X=[\textbf{x}_1, \textbf{x}_2,\cdots, \textbf{x}_n]\in \mathbb{R}^{m\times n}$ be a matrix or a dictionary consists of  $n$ known samples or atoms and $\textbf{y}\in \mathbb{R}^m$ is a sample to be represented, $m$ denotes the number of features to represent each sample.   We want to use $X$ to represent $\textbf{y}$ as:
  \begin{equation}
  \hat{\textbf{y}}=X\textbf{w}, \label{eq1}
  \end{equation} 
  where $\textbf{w} = [w_1, w_2, \cdots, w_n]^T\in \mathbb{R}^n$. In sparse learning, we want to obtain a solution of $\textbf{w}$ with majority of elements to be zero or close to zero. 
  A typical form of the objective function for sparse learning is given as:
   \begin{equation}
    \min_{\textbf{w}} f(X,\textbf{y},\textbf{w}) +\lambda \cdot \|\textbf{w}\|_0.  \label{eq2}
  \end{equation} 
    Very often, the first term,  also called the data term, is computed as 
  \begin{equation}
  f(X, \textbf{y}, \textbf{w})= \|\textbf{y}-X\textbf{w}\|^2_2 \label{eq3}
  \end{equation}
  In the data term, $\textbf{y}$ and $X$ can be the original pixel intensities or features computed from the pixel intensities. 
     The second term is the regularization term to ensure the sparsity and   $\lambda$ controls the weight of the term.
  Because of the  $\ell_0$ norm term, the minimization problem above is not convex and the solving of problem is NP-hard \cite{6873279}.  Very often, $\ell_1$ norm is used instead  to ensure sparsity  and convert the problem to a convex optimization problem \cite{CPA:CPA20132}. 
   
      One important issue in computing the sparse representation is to choose the atoms   to represent the data \cite{Wright10}. In another word, how to determine the non-zero elements and their corresponding weights in $\textbf{w}$. In signal processing,  the performances of sparse approximation algorithms often depend  on the mutual coherence of the data in the dictionary and much effort has been given to learn from incoherent dictionary  \cite{6451295}. 
    In computer vision, we often have to learn from overcomplete   dictionary  where the data is  not necessarily incoherent as well. 
  
  In the past, different variations and extensions of sparse learning  algorithms have been proposed, including $\ell_1$ norm, $\ell_1/\ell_q$ norm \cite{Argyriou2008},   fused lasso \cite{Tibshirani2005}, locality-constrained linear coding \cite{llc2010}, Laplacian sparse coding \cite{gao2010}, similarity weighted sparse representation \cite{guo2012}, sparse dissimilarity-constrained coding \cite{CJ15}, sparse group lasso \cite{sparsegrouplasso2010}, tree structured group lasso \cite{Jacob2009}, overlapping group lasso \cite{vijver2002}, ordered tree-nonnegative max-heap \cite{Zhao_thecomposite}, and trace-norm \cite{945730}.   
  The objective functions of these different sparse models can be  generally summarized as:
  \begin{equation}
     \min_\textbf{w} f(X, \textbf{y}, \textbf{w})+ \lambda \cdot r(\textbf{w}),  \label{eq4}
  \end{equation}
     where  $r(\textbf{w})$ is a function of $\textbf{w}$ only, such as $\ell_1$ norm $r(\textbf{w})=\|\textbf{w}\|_1$, $\ell_2$ norm $r(\textbf{w})=\|\textbf{w}\|_2^2$. In group lasso,  sparse group lasso, overlapping group lasso, the regularization term requires to divide $\textbf{w}$ into different overlapping or non-overlapping groups. The partition of   $\textbf{w}$ into overlapping or non-overlapping groups is often determined by properties of $X$.
     
     In this paper,
     we propose  a novel sparse range-constrained learning (SRCL)
     algorithm for medical image grading using a newly introduced range constraint regularization term. Different from most of existing
     sparse learning algorithms, SRCL integrates the final objective of grading the medical images  with the objective of finding a sparse representation into one function and solves the problem in one step. The method is able to find a sparse
     representation of the image based on images that are similar to the testing image in
     both the data or feature representation and the medical grading scores. 
      We then apply
     the new proposed SRCL to two different applications for CDR computation and cataract grading.
     %
     
    \textbf{Contribution:} Our main contributions are summarized as
    follows.  
  
     \begin{enumerate}
     	 \item We propose a new regularization term named range constraint to regularize the computation of the grades of medical images.
     	\item By combining the proposed range constraint with the objective of sparse learning, we propose a novel    SRCL method for medical image grading. It integrates the   objective of grading the medical images with the objective of finding a sparse representation into one function. 
     	 \item The proposed SRCL algorithm is applied to two different applications including CDR computation   and cataract grading. 
     	  Experimental results show that the proposed methods improve the accuracy in  CDR computation and cataract grading.
     	\item The method is a general approach and can be extended for other medical image grading applications.

     \end{enumerate}
     
     The rest of the paper is organized as follows. In section \ref{related}, we introduce the related work. In section \ref{SRL}, we introduce  the proposed  SRCL algorithm including the formulation of SRCL and its solution.  Section \ref{application}
     shows the use of SRCL for CDR computation  and cataract grading.
     Section \ref{Exp} show the experimental results, followed by the discussions and conclusions in the last section.
     
     \section{Related Work} \label{related}
   Recently, sparse learning methods have begun to be used for medical image grading, such as CDR computation from retinal fundus images \cite{YW13, CJ15, Cheng:17BOE} and cataract grading from slit-lamp lens images \cite{Xu13}.  These sparse learning methods compute the grade of medical images in two steps. The first step is to  reconstruct the testing  images based on a set of reference   images. The second step is to compute the grade based on the reconstruction coefficients. Here we briefly introduce these methods and discuss their limitations.
    \subsection{Reconstruction of images}
        Different constraints are used to regularize the selection of atoms from the overcomplete dictionary to reconstruct the testing data $\textbf{y}$ in the first step. 
    In \cite{YW13},   locality-constrained linear coding (LLC) \cite{llc2010} was used to reconstruct the disc image:
        \begin{equation}
    \min_{\textbf{w}} \|{\textbf{y}}-{X}\textbf{w}\|^2_2+\lambda_1  \|\textbf{c}\odot \textbf{w}\|_2^2, \label{llcobj} 
    \end{equation} 
    where $\textbf{c}=[c_1, c_2, \cdots, c_n]^T\in \mathbb{R}^n$ and $c_i= \exp(\frac{\|\textbf{y}-\textbf{x}_i\|^2}{2\sigma^2})$ denotes the Gaussian distance  between $\textbf{x}_i$ and $\textbf{y}$.  
    The limitation of the approach is that the solution is not sparse. In addition, the pixel-wise distance between two disc
    images suffers from various noise including blood vessels, disc alignment error, etc.
  This method has also been used to reconstruct cataract images for cataract grading \cite{yanwu2016}.
    
 In sparse coding (SC) \cite{sparsecoding}, $\ell_1$ norm regularization is added into the objective function to get a sparse solution:
       \begin{equation}
   \min_{\textbf{w}} \|{\textbf{y}}-{X}\textbf{w}\|^2_2+\lambda_1 \|\textbf{w}\|_1.  \label{scobj} 
   \end{equation} 
      However, the basic form of SC approach is not optimized for specific applications such as CDR estimation. 
      Inspired by the idea that discs with similar CDRs shall be similar in shape and the idea of a few similar discs shall be sufficient for CDR estimation,  sparse dissimilarity-constrained coding (SDC) has been proposed \cite{CJ15}:
      \begin{equation} \min_{\textbf{w}}\|{\textbf{y}}-{X}\textbf{w}\|^2_2+ \lambda_1 \|\textbf{w}\|_1+ \lambda_2\| \textbf{d}\odot \textbf{w}\|_2^2, 
    \end{equation} 
      where $\textbf{d}=[d_1, d_2, \cdots, d_n]^T\in \mathbb{R}^n$ and $d_i$ denotes the similarity based distance  between $\textbf{x}_i$ and $\textbf{y}$. 
    Compared with LLC, SDC includes the $\ell_1$ norm constraint to yield a sparse solution. In addition, a new distance $\textbf{d}$ based on  the similarity  of the discs is proposed. 
   SDC leads to sparse solution with higher weights in discs similar to the testing images, however, it may reconstruct the testing disc image using discs with quite different CDR values. This is not optimal for CDR computation in the next step. 
    
    In \cite{Cheng:17BOE}, the reference disc images have been divided into non-overlapping groups based on their CDR values  and  group sparsity has been included into the objective function:
    
      \begin{equation} \min_{\textbf{w}}\|{\textbf{y}}-{X}\textbf{w}\|^2_2+ \lambda_1 \|\textbf{w}\|_1+ \lambda_2\| \textbf{d}\odot \textbf{w}\|_2^2+\lambda_3   \sum_{i=1}^N\psi_i\|\textbf{w}_{G_i}\|_2,   
    \end{equation} 
   where $\textbf{w}$ is divided into $N$ non-overlapping groups $\textbf{w}_{G_1}, \textbf{w}_{G_1}, \cdots, \textbf{w}_{G_N}$, $\psi_i$ controls the weight of $i^{th}$ group,
   ${\lambda_3}$ controls the weight of the overall group lasso term.
   Although the sparse group sparsity leads to a sparse solution from each group, several groups with quite different CDRs may still be used. 
   
       \subsection{Grade computation}
   After solving $\textbf{w}$   based on different objective functions above,  the grade $\hat{g}$  of the testing image is then computed in the second step \cite{CJ15}:
  \begin{equation}
  \hat{g}=\frac{1}{\textbf{1}^T\textbf{w}}\textbf{w}^T\textbf{g} ,   \label{eq5}
  \end{equation} 
  where $\textbf{1}$ is a vector of 1s with length $n$, $\textbf{g}=[g_1, g_2, \cdots, g_n]^T\in \mathbb{R}^n$, $g_i$ denotes the grades for $\textbf{x}_i$. 
 
 It has been assumed that the minimization of the objective function in the first step will lead to a better estimation of the grade in the second step. However, this is not necessarily  true. 
         In fact, in the first step to find the representation of the data, the regularization constraint is computed from the data $\textbf{y}$ and the atoms in the dictionary $X$ only. The grade $\textbf{g}$ of the atoms in the dictionary $X$ has not been considered.  In another word,  the first step to reconstruct the images is independent from the second step to compute the grading score. The objective in the first step is not necessarily optimized for the second step.

\begin{figure}
    	\centering
    	{\subfigure[Image]{
    			\includegraphics[ width=1.5in]{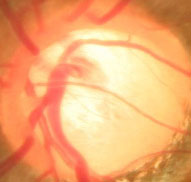}\label{fig1a}
    	}} 
    	{\subfigure[Manual Cup Boundary]{
    			\includegraphics[ width=1.5in]{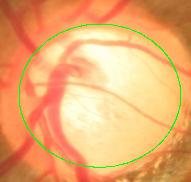}
    	}} 
    	\caption{An example of optic disc image: green line indicates the manual cup boundary, from which the CDR value of 0.79 is computed. The disc image is processed to be the input vector $\textbf{y}$ for sparse learning reconstruction. } \label{fig1}
    \end{figure}

\begin{figure}
     	\centering  {
     		\includegraphics[width=3.5in]{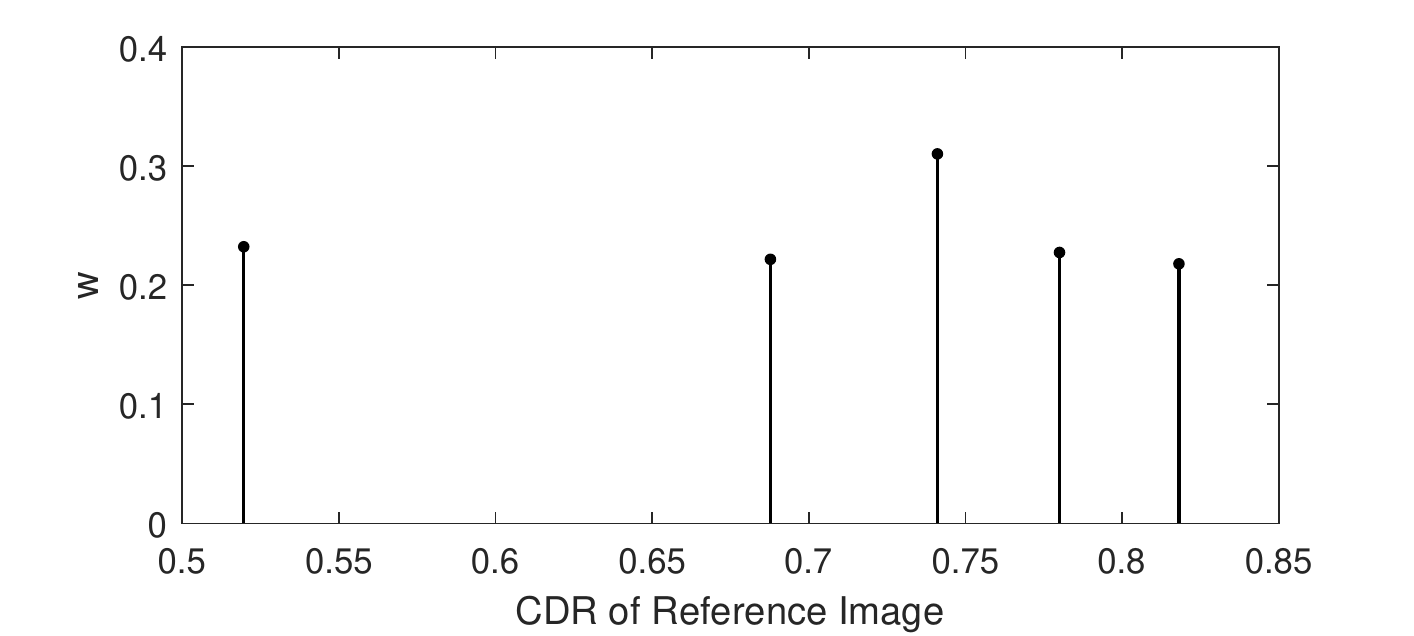}
     	}
     	\caption{Illustration of sparse coding in medical image grading: top five  coefficients in $\textbf{w}$ are plotted.  } \label{fig2}
     \end{figure}

    We use an actual  case  as an example and the SC \cite{sparsecoding} algorithm is used to reconstruct the retinal image in Fig. \ref{fig1a}. 
 In SC, we compute $\textbf{w}$ based on (\ref{scobj}).
 For simplicity, we plot the top five coefficients in $\textbf{w}$ and the CDRs of the corresponding discs. As shown in Fig. \ref{fig2}, the five discs corresponding to the top five coefficients have CDRs of 0.74, 0.51, 0.78, 0.68 and 0.81. 
  The second disc has a CDR of 0.51, much lower than other discs.   This   leads to challenges in estimate $\hat{g}$. 
  One would ask for the reason that a disc with obviously lower CDR of 0.51 can have a high weight in $\textbf{w}$. This is actually because the data term is not perfect and it is corrupted by many factors, e.g., retinal vessels, error in disc segmentation, un-even illumination across the disc, etc.   

     It should be noted that the above bias might also be caused by the linear image grading computation in $(\ref{eq5})$, which   is actually an approximation of actual grading score. 
To explain this, we  artificially create a disc  image  $\hat{\textbf{y}}=\frac{1}{2}(\textbf{x}_1+\textbf{x}_2)$ and its ideal reconstruction  $\hat{\textbf{w}}$ with two non-zero values $\hat{w}_1=\hat{w}_2=0.5$ and   $\hat{w}_i =0, \forall i\geq 3$. Based on equation (\ref{eq5}), the computed CDR would be $\frac{1}{2}(g_1+g_2)$. However, the manual CDR of $\hat{\textbf{y}}$ depends on the summation of $\textbf{x}_1$ and $\textbf{x}_2$ and may not be $\frac{1}{2}(g_1+g_2)$.  The difference in between  can be large if  $|g_1-g_2|$ is large.


\section{Sparse Range-constrained Learning}\label{SRL}


Since the objective of our reconstruction is to compute the grading of $\textbf{y}$, intuitively, we intend to have high weights for  atoms with similar grades and low or even zero weights for those with quite different grades. Ideally, we want to find a subset $\phi$ of the atoms whose grades satisfy:
\begin{equation}
   |\hat{g}-g_m| < \epsilon,  \forall \textbf{x}_m \in \phi
\end{equation}
where $\hat{g}$ denotes the grade of $\textbf{y}$, $g_m$ denotes the grade of $\textbf{x}_m$ in $\phi$ and $\epsilon$ is a small positive value. 
%
\subsection{Range constraint}
Inspired from the above discussions, we propose a new constraint to use
atoms that have grading scores close to the testing image. In another word, we want to find a
 subset of atoms $\{\textbf{x}_i \}, i\in \{1, 2, \cdots, n\}$, whose grades $g_i$ are close to that of the
 testing image. Therefore, we want to minimize $ |\hat{g}-g_i|$ for entries
 with non-zero weights $w_i$.   The proposed  range constraint (RC) term is defined as:
 \begin{equation}
    \sum_i (w_i(\hat{g}-g_i))^2
 \end{equation}
We name the proposed constraint term as range constraint because it tends to reduce the range  
 $\max_{p,q}|g_p-g_q|$  with non-zero $w_p$ and $w_q$.  
 The matrix form of the RC term is given as
 \begin{equation}
    \|\textbf{w}\odot (\hat{\textbf{g}} -\textbf{g})  \|_2^2,
 \end{equation}
 where $\hat{\textbf{g}}$ is a vector of $\hat{g}$ with length $n$.
  Intuitively, an atom $\textbf{x}_i$ shall have a low or even zero weight $w_i$ if    $|\hat{g}-g_i|$ is large.  
 Different from previous regularization terms which are   functions of $\textbf{w}$, RC is a function of both $\textbf{w}$ and $\hat{\textbf{g}}$.

 \subsection{Formulation of SRCL}
 As we mentioned earlier, a limitation of previous sparse learning based algorithms for medical image grading is that the step of   reconstructing the testing images from the reference is independent from the step of computing the grades.  To overcome this limitation,  we propose to integrate  the above RC  term with the data term
 $\|\textbf{y}-X\textbf{w}\|^2_2$  and the regularization term $r(\textbf{w})$. The  general objective function for SRCL is given by:
  \begin{equation}
  \min_{\textbf{w},\hat{\textbf{g}}} \|\textbf{y}-X\textbf{w}\|^2_2+ \lambda \cdot r(\textbf{w}) +\gamma  \cdot \|\textbf{w}\odot (\hat{\textbf{g}} -\textbf{g})  \|_2^2, \label{srlobj}
  \end{equation}
  where $\gamma$ controls the weight
  of RC term.
  
  Different from previous methods \cite{sparsecoding}, \cite{CJ15}, \cite{sparseGroupLasso}, SRCL considers both $\textbf{w}$ and $\hat{\textbf{g}}$ in the objective function.  This function is general and can be combined with different forms of  $r(\textbf{w})$. The benefits of new regularizer are twofold. On the one hand,
  by integrating this term into the objective function, we can
  solve $\textbf{w}$ and ${\hat{\textbf{g}}}$ simutaneously in one step. On the second hand,
  it also affects the solution of $\textbf{w}$. More specifically, the new
  term leads to a solution w such that wi is non-zero only for
  a small number of atoms with very close grades. This helps
  to reduce the bias as we have discussed in Section II.  
  Below, we show how to solve $\textbf{w}$ and $\hat{{\textbf{g}}}$.
  
   \subsection{Solution of SRCL} \label{solution}
  
   Since we have two unknowns $\textbf{w}$ and $\hat{\textbf{g}}$ in (\ref{srlobj}), we solve them by alternatively solving the two subproblems below until convergence:
 %
%
   \begin{equation}
     \min_{\textbf{w}_t} \|\textbf{y}-X\textbf{w}_t\|^2_2+  \lambda\cdot r(\textbf{w}_t) +\gamma  \cdot  \|\textbf{w}_t\odot (\hat{\textbf{g}}_{t-1} -\textbf{g})  \|_2^2,  \label{sub1}
   \end{equation} 
     \begin{equation}
       \min_{\hat{\textbf{g}}_t} \|\textbf{y}-X\textbf{w}_t\|^2_2+  \lambda \cdot r(\textbf{w}_t) +\gamma\cdot  \|\textbf{w}_t\odot (\hat{\textbf{g}}_t -\textbf{g})  \|_2^2.\label{sub2}
 \end{equation} 
   To solve the first subproblem in (\ref{sub1}),  we rewrite  it as:
   \begin{align} &\min_{\textbf{w}_t} \|\textbf{y}-X\textbf{w}_t\|^2_2+  \lambda \cdot r(\textbf{w}_t) +\gamma  \cdot  \|\textbf{w}_t\odot (\hat{\textbf{g}}_{t-1} -\textbf{g})  \|_2^2\nonumber \\
   & = \min_{\textbf{w}_t} \left\|\left[\begin{array}{c}\textbf{y}  \nonumber \\ \textbf{0}
   \end{array}\right]-
   \left[\begin{array}{c}X  \nonumber \\
   \sqrt{\gamma}\Delta   \\
   \end{array}\right]\textbf{w}_t \right\|^2_2 + \lambda\cdot r(\textbf{w}_t)\nonumber \\
   &  =  \min_{\textbf{w}_t}   \|\hat{\textbf{y}}-\hat{X}\textbf{w}_t\|^2_2+\lambda\cdot r(\textbf{w}_t),        \label{eq09}
   \end{align}
   where  $\hat{\textbf{y}}=\left[\begin{array}{c}\textbf{y} \\
   \textbf{0}    
   \end{array}\right] \in \mathbb{R}^{m+n}$, $\textbf{0}$ is a vector of 0s with length $n$,  $\hat{X}=\left[\begin{array}{c}X \\
   \sqrt{\gamma}\Delta   \end{array}\right]  \in \mathbb{R}^{(m+n)\times n}$ and  $\Delta \in \mathbb{R}^{n\times n}$ denotes a diagonal matrix with $\Delta(i,i)=\hat{g}_{t-1}-g_i$.
   This implies that we can always integrate the RC with the data term and  solve the subproblem in (\ref{sub1}) as the original problem without the range constraint.  
 
  The second subproblem in (\ref{sub2}) has a closed-form solution. Since $\hat{\textbf{g}}_t$ is a vector of $\hat{g}_t$ with length $n$, we compute the first derivative of the term with respect to   $\hat{g}_{t}$.  
  By letting the first derivative of the term to be zero, it is easy to find out  that 
  \begin{equation}
   \hat{g}_{t}=\frac{\sum_{i=1}^n w_i^2g_i}{\sum_{i=1}^n w_i^2}
  \end{equation}
  We iteratively solve the above two subproblems until convergence or the maximum number of iteration arrives. 
  
  It can be seen that:
  \begin{equation}
   \hat{g}_{t}=\frac{\sum_{i=1}^n w_i^2g_i}{\sum_{i=1}^n w_i^2}\leq=\frac{\sum_{i=1}^n w_i^2g_{max}}{\sum_{i=1}^n w_i^2}=g_{max},
  \end{equation}
  where $g_{max}=max\{g_1, g_2, \cdots, g_n \}$ denotes the maximum value of $g_i, i=1, 2, \cdots, n$.
  
  Similarly, it can also be proved that $\hat{g}_t \geq g_{min}$ and $g_{min}$ denotes the minimum value of $g_i, i=1, 2, \cdots, n$. Since $g_{min}\leq \hat{g}_t \leq g_{max}$ in each iteration, the final computed grades will also 
  be in the range of $[g_{min}~g_{max}]$. Therefore, the computed $\hat{g}$ is always in the same scale range of that for the dictionary images.
  
  Depending on the regularization term $r(\textbf{w})$, we may solve the subproblem in (\ref{sub1}) differently.
  Below, we discuss the three different forms of SRCL algorithms based on different $r(\textbf{w})$ from SC, SDC and SSGL. By combining the RC with these  methods, we obtain three different SRCL methods in this paper, denoted as SC+RC, SDC+RC and SSGL+RC.

 \subsubsection{SC+RC}

  \begin{equation}
 r(\textbf{w})=\frac{\lambda_1}{\lambda}\|\textbf{w}\|_1.
 \end{equation} 
 In this situation, we use the objective function $r(\textbf{w})$ as that of SC \cite{sparsecoding}: 
 \begin{align}
  f(\textbf{w}_t)&= \|\textbf{y}-X\textbf{w}_t\|^2_2+  \lambda_1 \cdot \|\textbf{w}_t\|_1 +\gamma    \|\textbf{w}_t\odot (\hat{\textbf{g}}_{t-1} -\textbf{g})  \|_2^2\nonumber \\
  &=\|\hat{\textbf{y}}-\hat{X}\textbf{w}_t\|^2_2+\lambda_1\cdot \|\textbf{w}_t\|_1. \label{016}
  \end{align}
 Since $f(\textbf{w}_t)$ integrates SC with RC, we denote the  approach as ``SC+RC".
 Minimization of (\ref{016}) is a standard $\ell_1$-norm regularized least square minimization problem.
It has been shown that this
unconstrained convex optimization problem can be solved by least
angle regression (LARS) \cite{lars2004}.

  \subsubsection{SDC+RC}
  	
  \begin{equation}
 r(\textbf{w})=\frac{\lambda_1}{\lambda}\|\textbf{w}\|_1+\frac{\lambda_2}{\lambda}\| \textbf{d}\odot \textbf{w}\|_2^2.
  \end{equation}
  In this situation, we use the objective function $r(\textbf{w})$ as that of 	SDC \cite{CJ15}. Therefore we denote this approach as ``SDC+RC".
   It shall be noted that the   `distance' here can be computed by any form, such as dissimilarity or other distance between  $\textbf{x}_i$  and $\textbf{y}$.  We have:
   \begin{equation} f(\textbf{w}_t)=\|\hat{\textbf{y}}-\hat{X}\textbf{w}_t\|^2_2+ \lambda_1 \|\textbf{w}_t\|_1+ \lambda_2\| \textbf{d}\odot \textbf{w}_t\|_2^2. \label{eq018}
  \end{equation} 
    It has been shown in  \cite{CJ15} that the third term in (\ref{eq018}) can be merged into the first term. Therefore,  ``SDC+RC"  is still a $\ell_1$-norm regularized least square minimization problem that can be solved by LARS. 

%
  \subsubsection{SSGL+RC}
 \begin{equation} r(\textbf{w})=\frac{\lambda_1}{\lambda} \cdot
  \|\textbf{w}\|_1+\frac{\lambda_2}{\lambda}\|\textbf{d} \odot \textbf{w} \|_2^2+ \frac{\lambda_3}{\lambda} \cdot  \sum_{i=1}^N\psi_i\|\textbf{w}_{G_i}\|_2. 
  \end{equation}
   In this situation, we use the objective function $r(\textbf{w})$ as that of SSGL \cite{Cheng:17BOE} and the approach is therefore denoted as ``SSGL+RC".
  Similarly, we have:
   \begin{align} f(\textbf{w}_t)&=\|\hat{\textbf{y}}-\hat{X}\textbf{w}_t\|^2_2+ \lambda_1 \|\textbf{w}_t\|_1+ \lambda_2\| \textbf{d}\odot \textbf{w}_t\|_2^2 \nonumber
   \\ &+\lambda_3   \sum_{i=1}^N\psi_i\|\textbf{w}_{t_{G_i}}\|_2.  \label{eq019}
  \end{align} 
  As illustrated in \cite{Cheng:17BOE}, minimizing the objective function in (\ref{eq019}) can be converted to a standard sparse group lasso problem. We solve the
  problem using the sparse learning with efficient projection (SLEP) package \cite{Liu:2009:SLEP:manual}.
 
 In this paper, we mainly discuss the above three forms of $r(\textbf{w})$.   We leave it to the respective readers to find the applications of other  regularization terms.

\section{SRCL for Medical Image Grading} \label{application}
In this section, we show the use of SRCL for two different medical image grading applications: CDR computation from retinal fundus images and cataract grading from slit-lamp lens images.
\subsection{Cup-to-disc ratio computation}
The first application is to compute the CDR from retinal fundus images. CDR is often computed as the ratio of vertical cup diameter to vertical disc diameter.
 It is a commonly used indicator in glaucoma detection. 
  A higher CDR often indicates a higher risk of glaucoma. Previously, 
SC \cite{sparsecoding}, SDC \cite{CJ15}  and SSGL\cite{Cheng:17BOE} have been proposed for CDR computation. In SDC and SSGL,  the distance   $\textbf{d}$ is computed  based on  the shape similarity of the disc images as in \cite{CJ15}.  
 
   In this paper, we use all the three previous approaches SC, SDC and SSGL as baselines to show how the RC term improves the CDR computation. As showed in Section \ref{solution}, by adding the RC term into the objective function of SC, SDC and SSGL,  we have three new algorithms  denoted as SC+RC, SDC+RC and SSGL+RC, respectively.   Recall that in Section \ref{solution}, we solve $\textbf{w}_t$ and $\hat{g}_t$ iteratively. We need to obtain an initial grade $\hat{g}_0$ to compute $\textbf{w}_1$. For SC+RC, SDC+RC and SSGL+RC, we use the results from SC, SDC and SSGL as the initial values for $\hat{g}_0$, respectively. 
    CDR is obtained after  solving the problems in (\ref{sub1}) and (\ref{sub2}) iteratively.
%
%
%

\subsection{Cataract grading}
 Cataracts are the leading cause of   blindness worldwide. Cataract grading  is essential for diagnosis and progression monitoring of the disease. In cataract grading, the images are first graded with decimal grading scores that
 range from 0.3  to 5.0 by professional graders based on
 the Wisconsin protocol \cite{Klein1990}.   The protocol takes the ceiling of each decimal grading score as the integral grading score.  
 In our grading, we use the decimal grading scores before the ceiling as the ground truth scores. 
 Similar to CDR computation, we also use SC, SDC and SSGL as the baselines.
 Therefore we have SC+RC, SDC+RC and SSGL+RC for cataract grading as well.
 Following the method in \cite{Xu13}, the distance $d_i$ is computed as $ \chi^2$-distance between the testing image $\textbf{y}$ and the reference image $\textbf{x}_i$ for SDC, SSGL, SDC+RC and SSGL+RC.
  Similar to that in CDR computation, we also use SC, SDC and SSGL to get initial grades for SC+RC, SDC+RC and SSGL+RC, respectively.
  
A decimal score is obtained for cataract grading by solving the problems in (\ref{sub1}) and (\ref{sub2}) iteratively. A ceiling operation is further applied to get the integral grading score.

\section{Experimental
Results} \label{Exp}
\subsection{Cup-to-disc ratio computation}

\subsubsection{Data Set}
We use 650 images from ORIGA data set \cite{origa}. The data set
consists of 168 images from glaucomatous eyes and 482 images from
randomly selected normal eyes from a population based study.
The ground truth CDR is computed as the ratio of vertical cup diameter to vertical disc diameter from optic cup and optic disc boundaries manually labelled by trained professionals. 
The images are partitioned into two subsets $A$ and $B$ previously and we followed the same partition in this paper \cite{Cheng:17BOE}. In our experiments, we use images from set $A$ as reference or training images and images from set $B$ as testing images.
  Since different disc images have different dimensions, we resize all the images to  the same size. Following that in \cite{CJ15}, we resize all disc images to $50\times 50$ pixels for less computational cost. 
%
 \subsubsection{Evaluation Metrics}
The following criteria are used in our evaluation:
 \begin{itemize} 
 	 \item  mean absolute CDR error: \begin{equation} \delta=\frac{1}{N}\sum_i^N |GT_{i}-CDR_i|,
 \end{equation} where
$GT_{i}$ and $CDR_i$ denote the manually and automatically measured CDR of sample
$i$, $N$ denotes the number of testing images.
 \item The correlation:  the Pearson's correlation
coefficient between the manually measured CDRs and automatically computed CDRs is computed.
  \end{itemize}
 \subsubsection{Results}
\begin{figure}
 	\centering
 	{\subfigure[SC]{
 			\includegraphics[width=3.0in, height=1.2in]{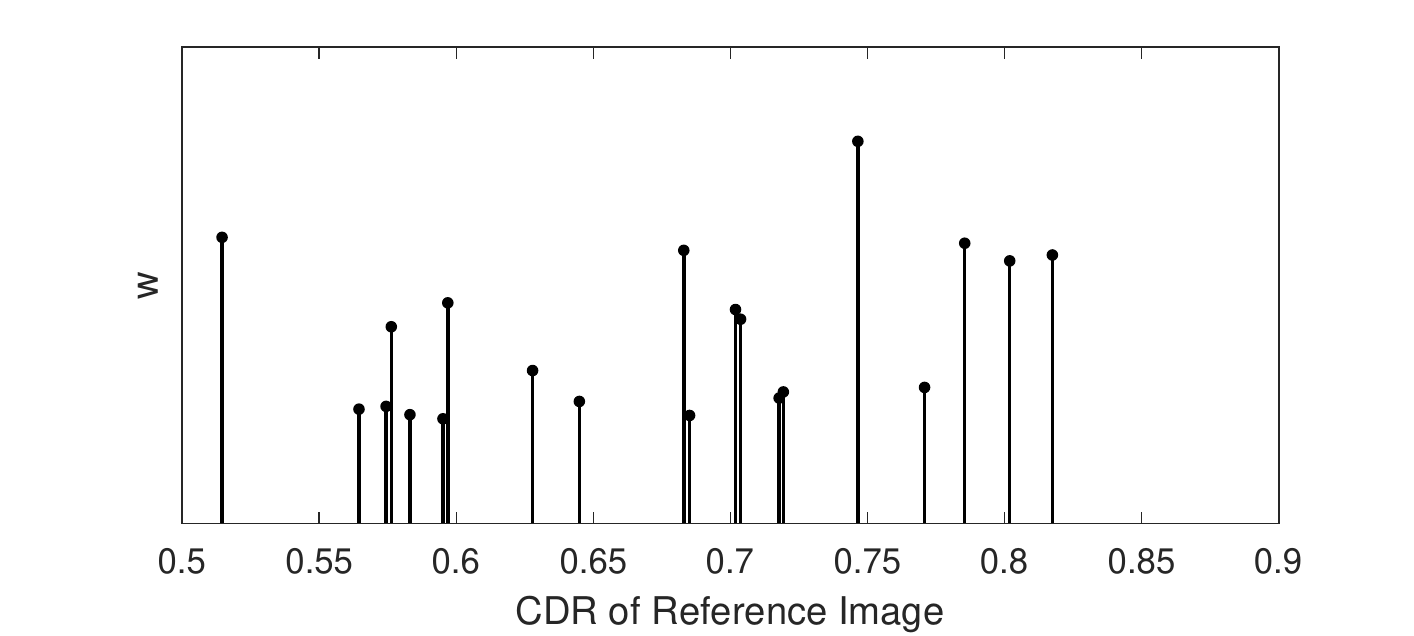}
 	}}
 	{\subfigure[After one iteration in SC+RC]{
 			\includegraphics[width=3.0in, height=1.2in]{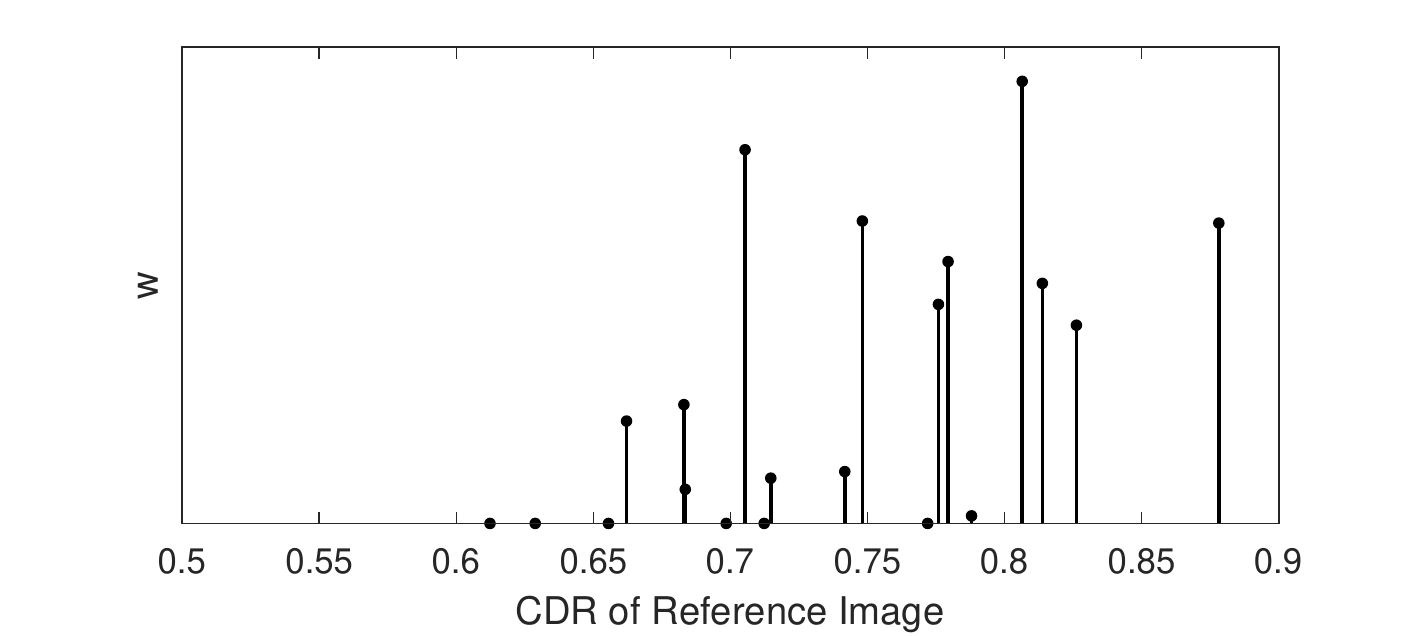}
 	}}
 
 	\caption{Top 20 non-zero elements in $\textbf{w}$ when reconstruct an image using SC and SC+RC.  } \label{figlambda1}
 \end{figure}

 To show how the RC term affects the results, we first look at how the parameter $\textbf{w}_t$ changes as $t$ iterates. We take the SC+RC as an example. 
 Fig. \ref{figlambda1} shows the top 20 non-zero elements in $\textbf{w}_t$ before and after the first iteration   for the disc in Fig. \ref{fig1} with manual CDR of 0.79. In the figure,   $x$-axis indicates the CDR value of the images with non-zeros weights from the dictionary $X$ and $y$-axis indicates the corresponding weight in $\textbf{w}_t$ for these images. From the figure, we can see that the elements with top 20 non-zero weights are have a wide range of CDR values from 0.51 to 0.82. After one iteration, the elements with top 20 non-zero weights have a smaller range of CDR values from  0.67 to 0.86.  
  Correspondingly, the computed CDR values before and after the iteration are 0.68 and 0.77  respectively. Further iterations  continue  to reduce the range.

 \begin{figure}
 	\centering
 	{\subfigure[CDR Error]{
 			\includegraphics[width=3.0in]{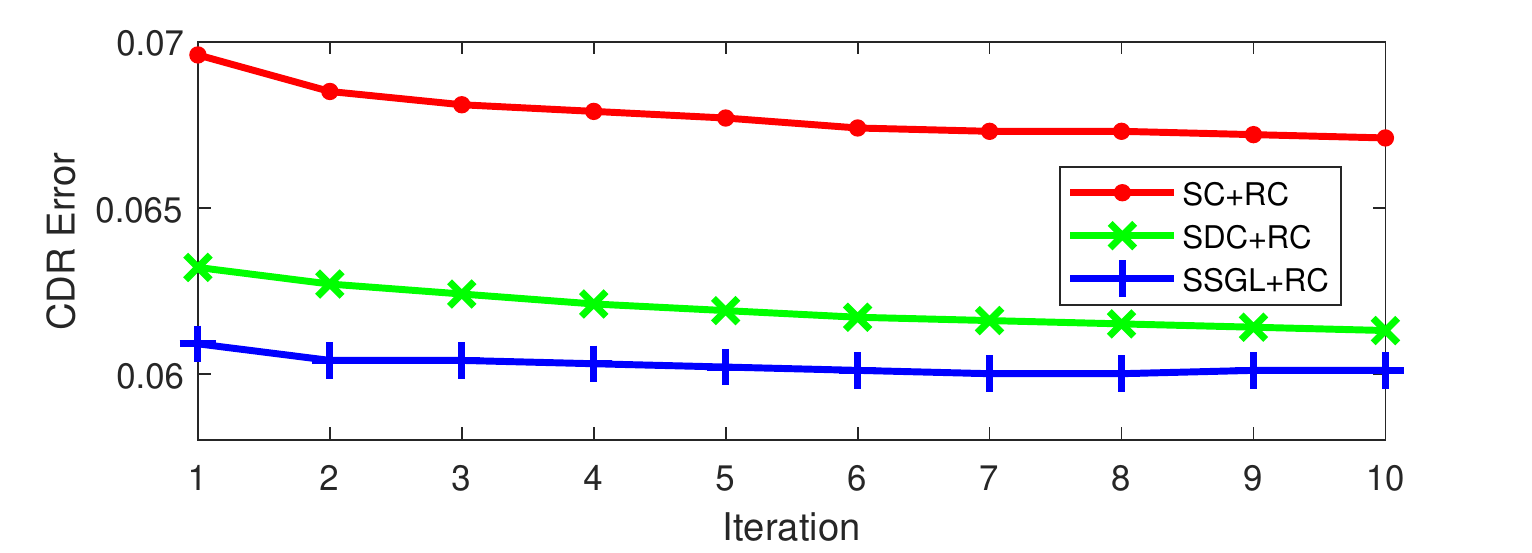}
 	}}
 	{\subfigure[Correlation]{
 			\includegraphics[width=3.0in]{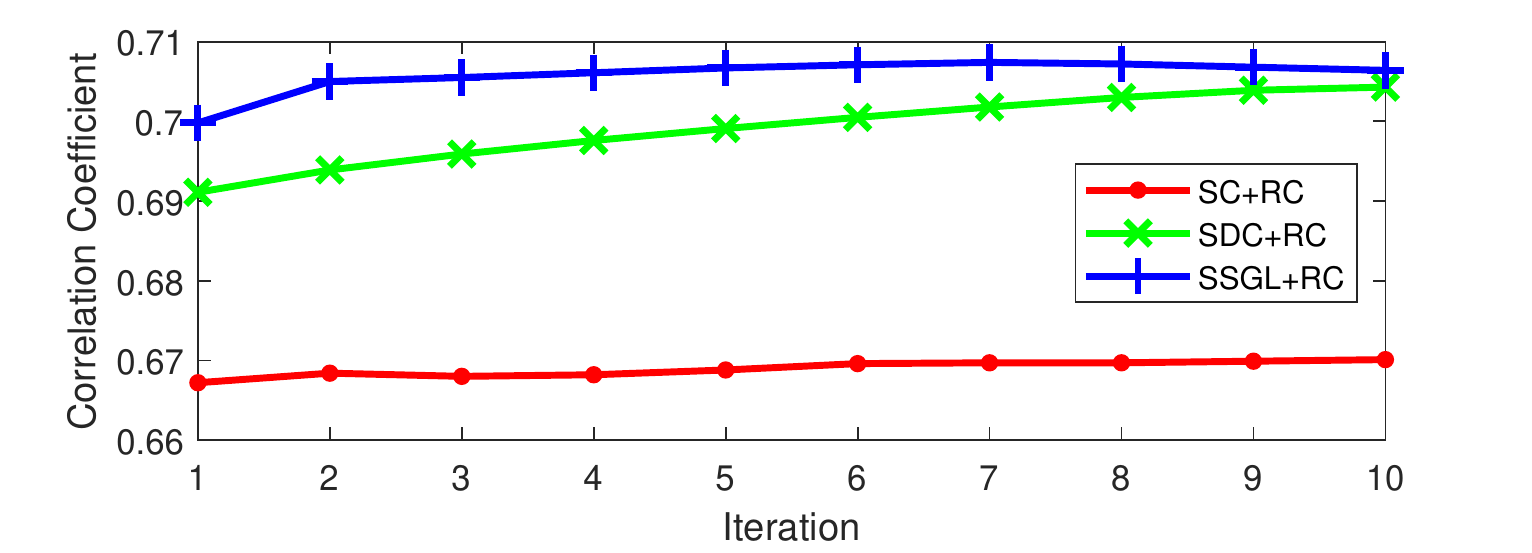}
 	}}

 	\caption{Performance changes during the iteration of SC+RC, SDC+RC and SSGL+RC. } \label{fig4}
 \end{figure}

\begin{figure*}
	\centering
	{\subfigure[Sample 1: manual CDR 0.671]{
			\includegraphics[height=1.2in,width=1.2in]{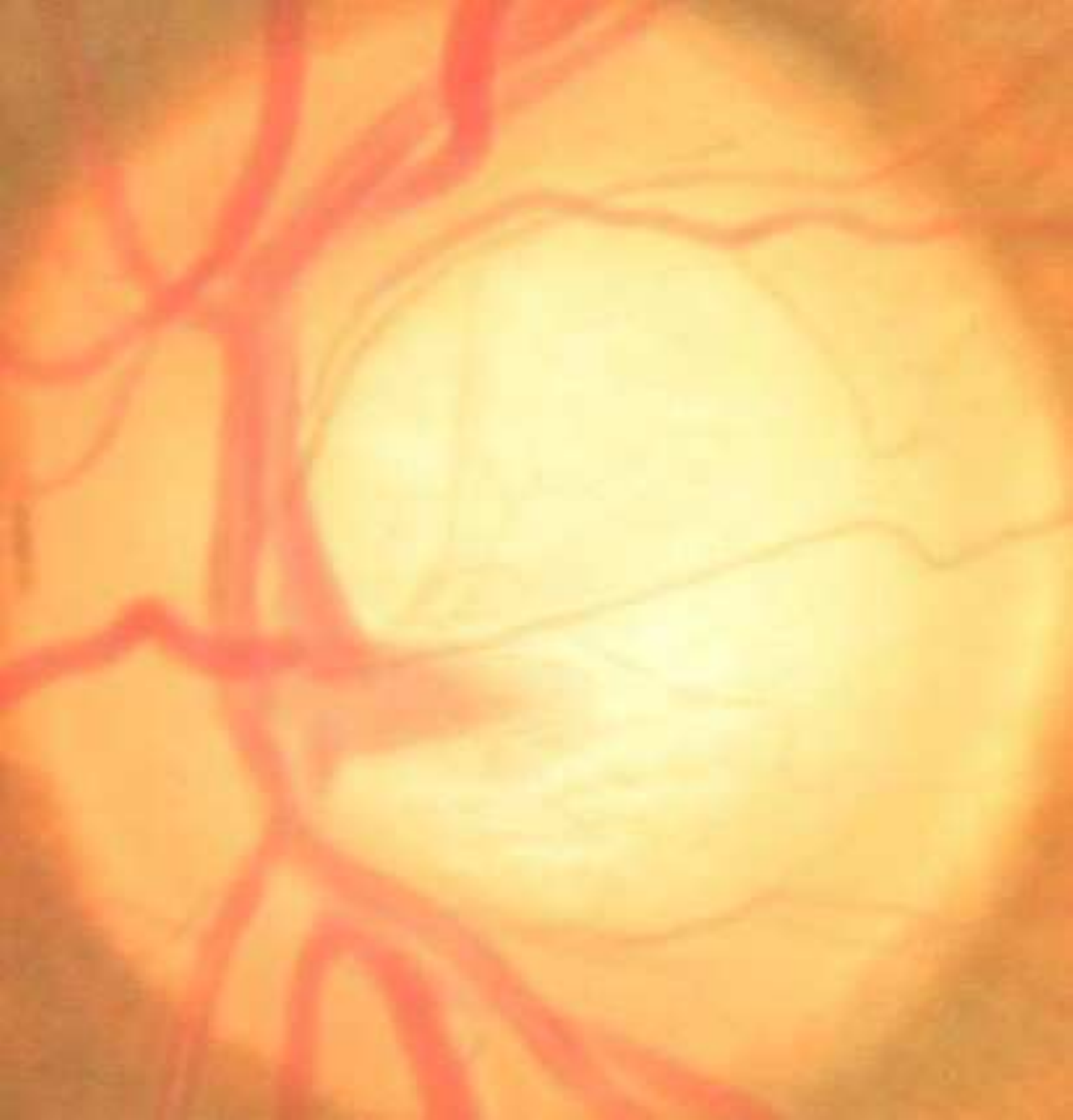}
	}}
	{\subfigure[Reconstruction coefficients $\textbf{w}$ by SDC and estimated CDR 0.639]{
			\includegraphics[height=1.2in]{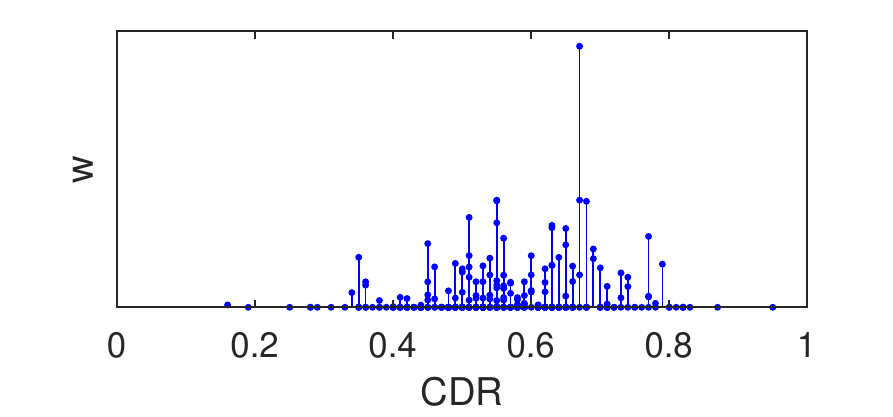}
	}}
	{\subfigure[Reconstruction coefficients $\textbf{w}$ by SDC+RC and estimated CDR 0.670]{
			\includegraphics[height=1.2in]{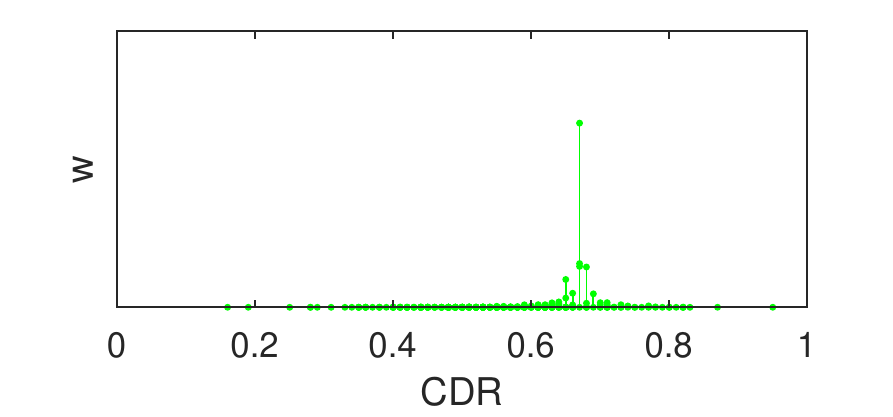}
	}}
	{\subfigure[Sample 2: manual CDR 0.619]{
			\includegraphics[height=1.2in,width=1.2in]{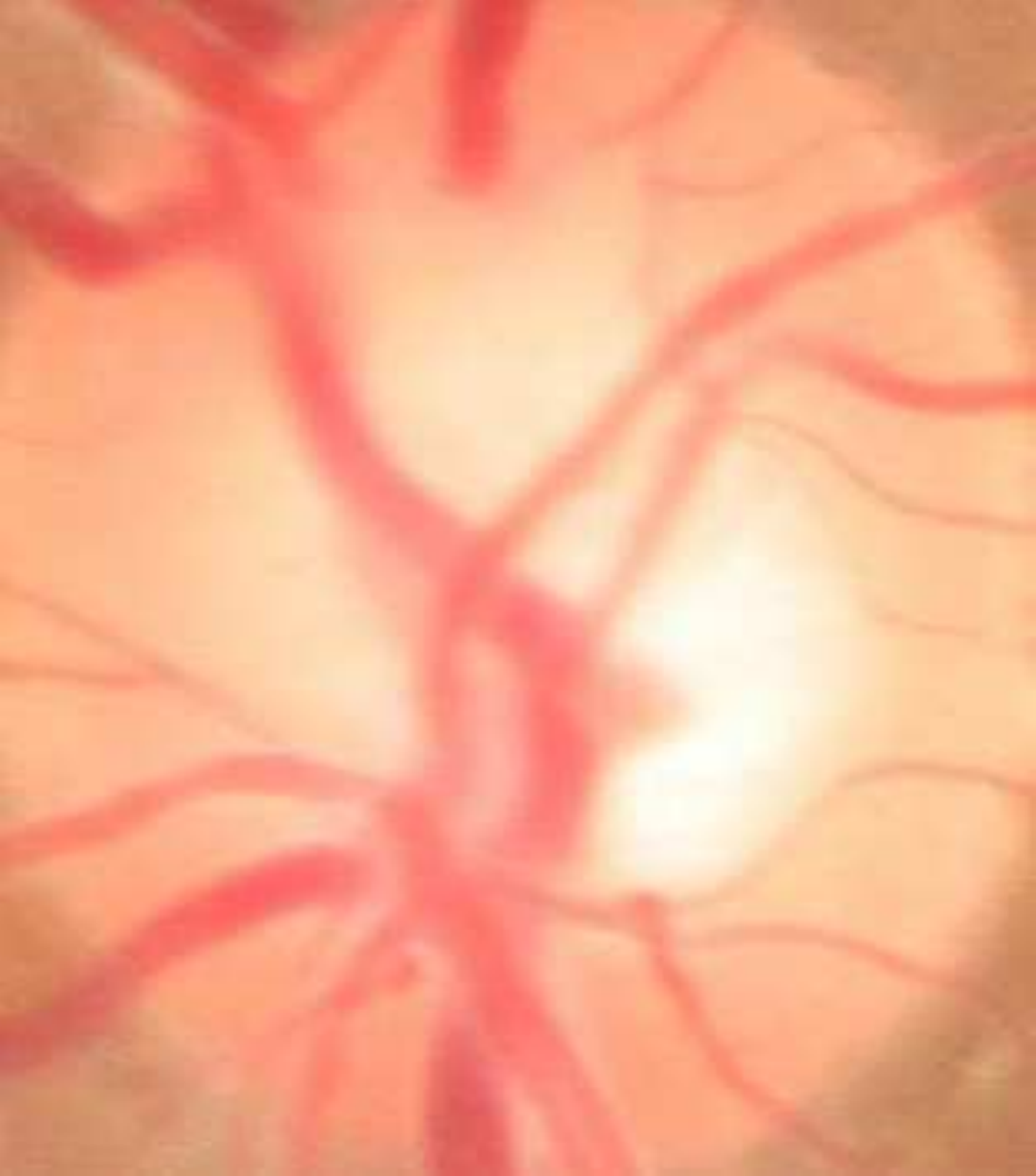}
	}}
	{\subfigure[Reconstruction coefficients $\textbf{w}$ by SDC and estimated CDR 0.523]{
			\includegraphics[height=1.2in]{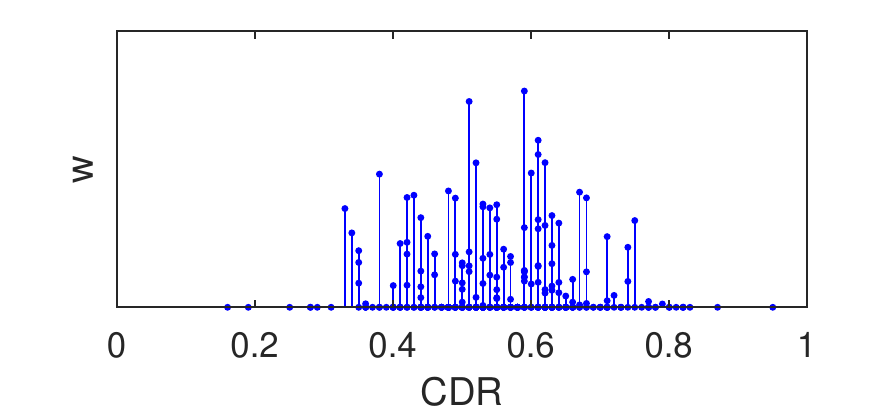}
	}}
	{\subfigure[Reconstruction coefficients $\textbf{w}$ by SDC+RC and estimated CDR 0.559]{
			\includegraphics[height=1.2in]{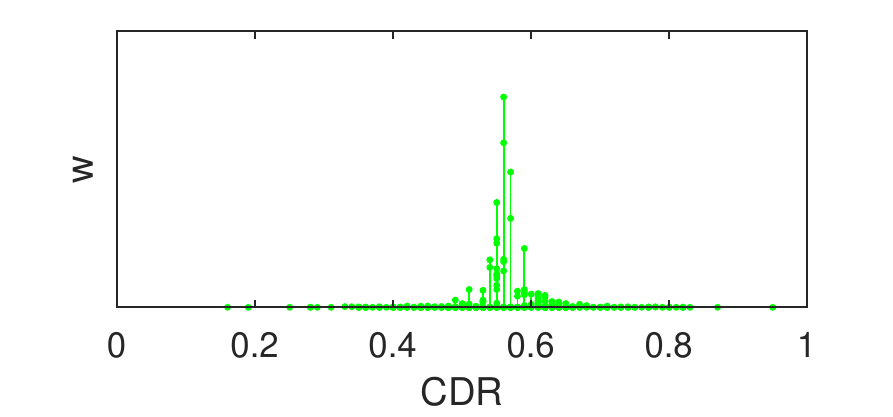}
	}}

	\caption{Samples images with reconstruction coefficients $\textbf{w}$ and resultant CDRs by SDC and SDC+RC methods.} \label{fig44}
\end{figure*}

\begin{figure*}
	\centering
	{\subfigure[Sample 1: manual CDR 0.497]{
			\includegraphics[height=1.2in,width=1.2in]{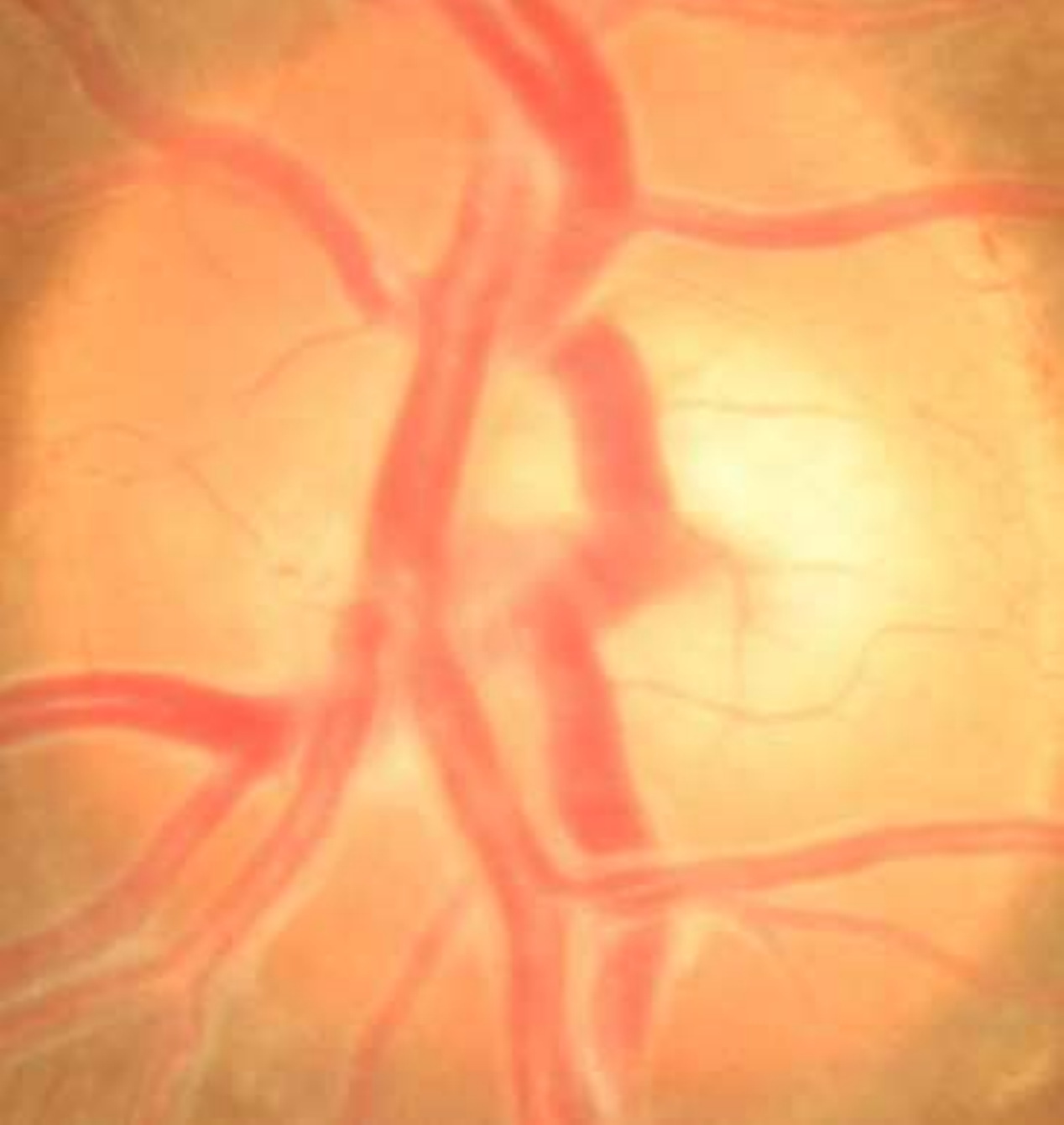}
	}}
	{\subfigure[Reconstruction coefficients $\textbf{w}$ by SSGL and estimated CDR 0.563]{
			\includegraphics[height=1.2in]{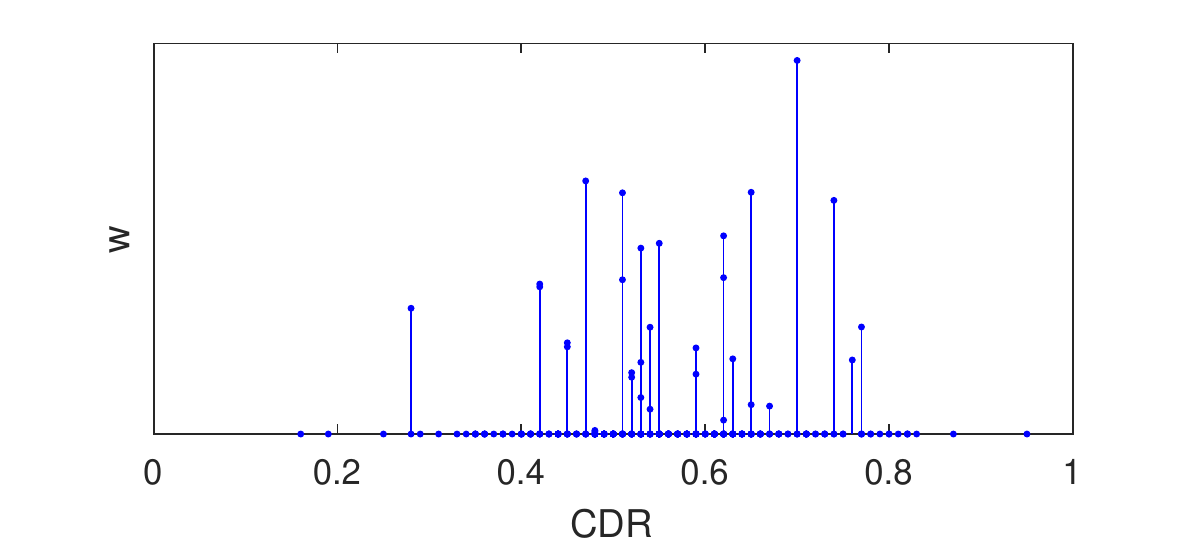}
	}}
	{\subfigure[Reconstruction coefficients $\textbf{w}$ by SSGL+RC and estimated CDR 0.531]{
			\includegraphics[height=1.2in]{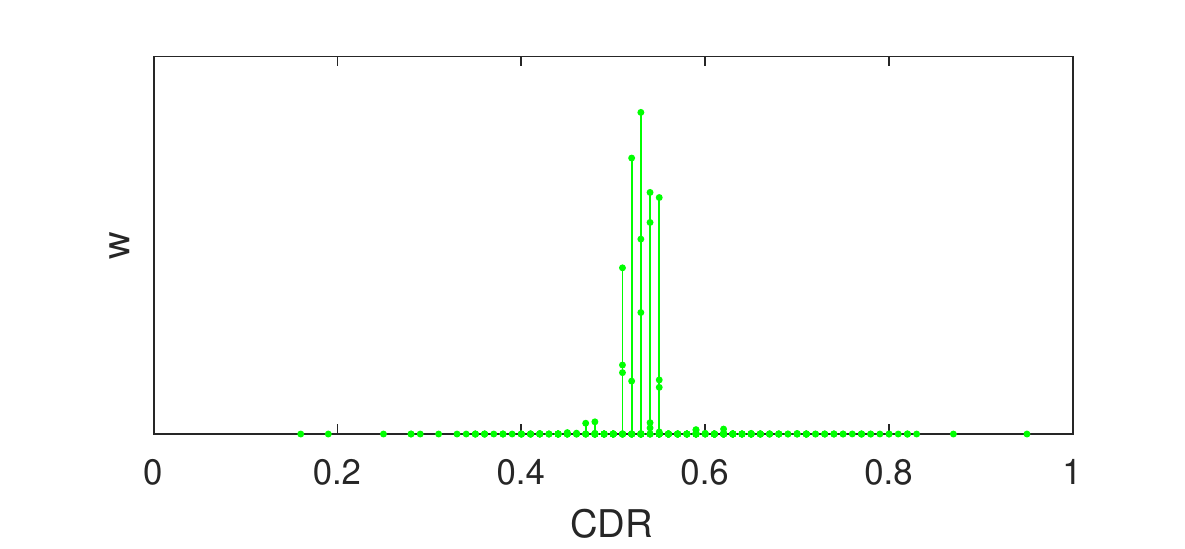}
	}}
	{\subfigure[Sample 2: manual CDR 0.591]{
			\includegraphics[height=1.2in,width=1.2in]{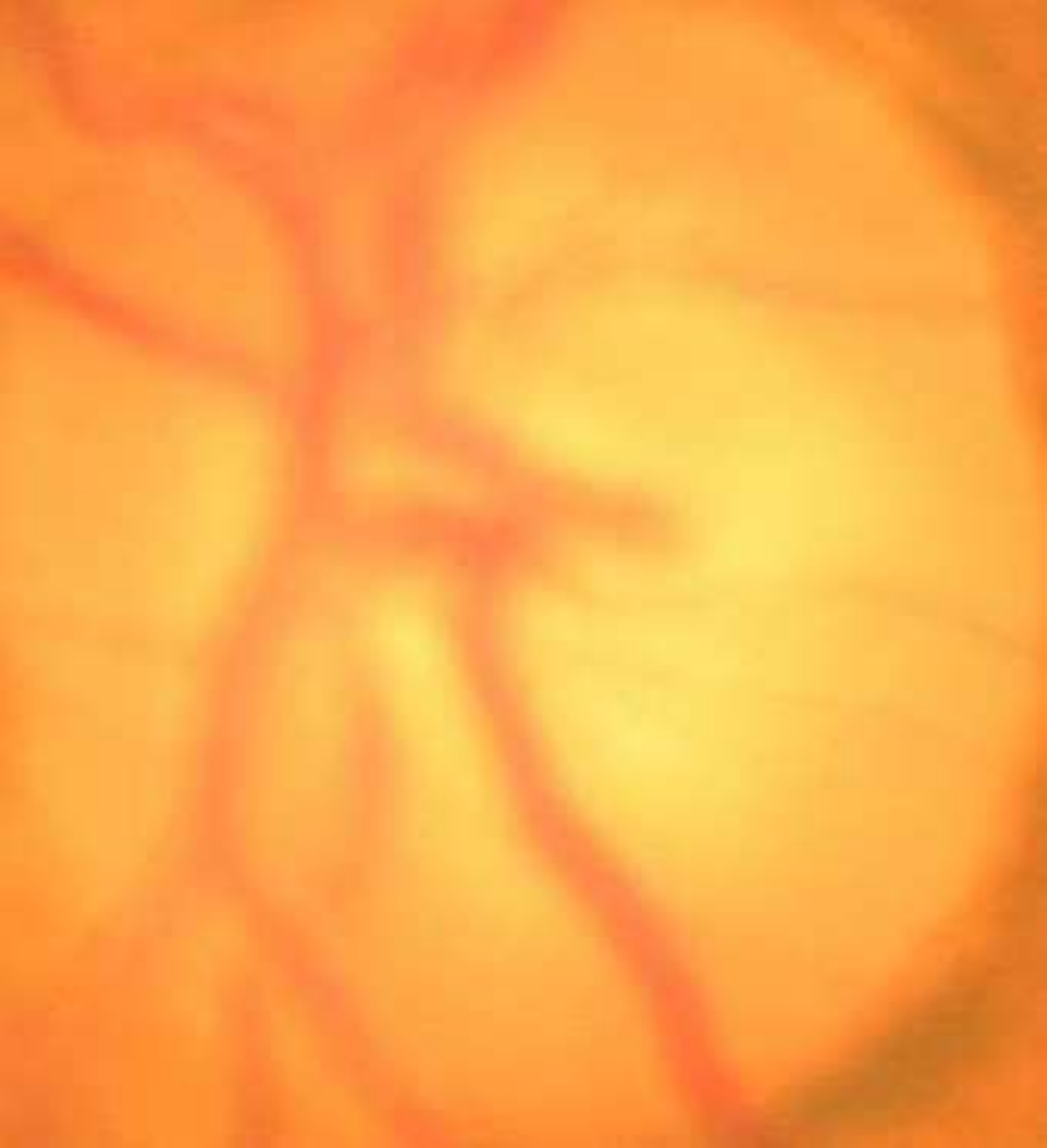}
	}}
	{\subfigure[Reconstruction coefficients $\textbf{w}$ by SSGL and estimated CDR 0.630]{
			\includegraphics[height=1.2in]{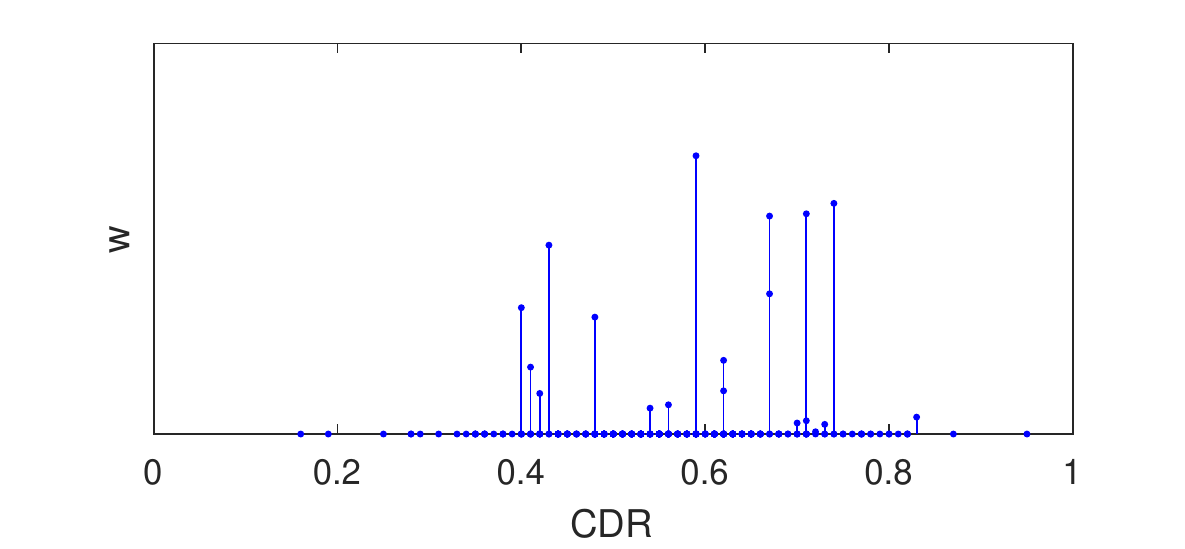}
	}}
	{\subfigure[Reconstruction coefficients $\textbf{w}$ by SSGL+RC and estimated CDR 0.608]{
			\includegraphics[height=1.2in]{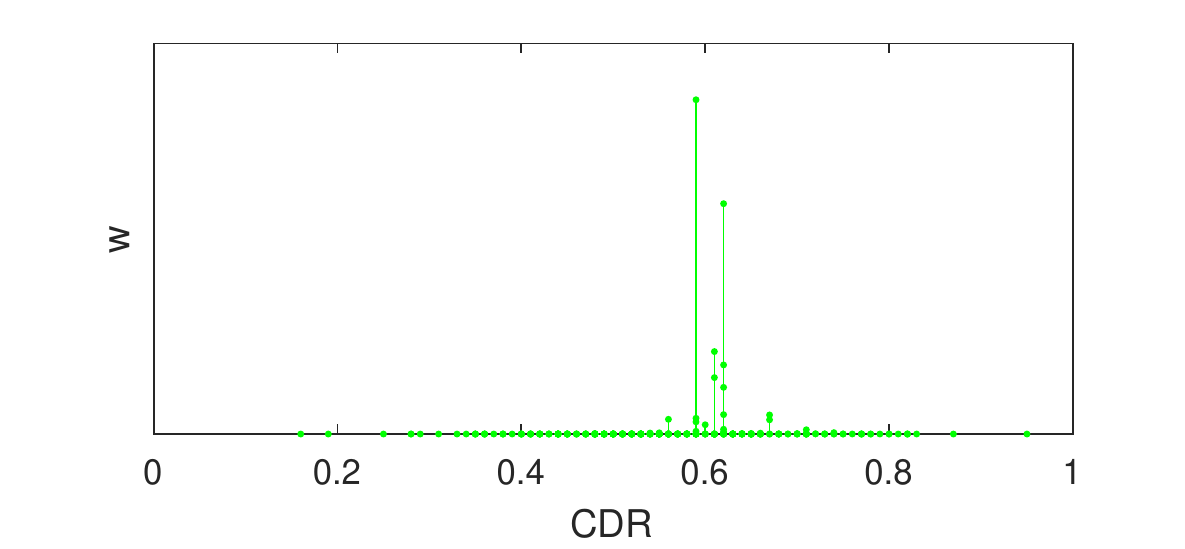}
	}}
	\caption{Samples images with reconstruction coefficients $\textbf{w}$ and resultant CDRs by SSGL and SSGL+RC methods.} \label{fig45}
\end{figure*}

 Fig. \ref{fig4} shows the mean CDR errors and the correlation coefficients with respect to manual ground truth by  SC+RC, SDC+RC and SSGL+RC for $t$ from 1 to 10 computed from all 325 testing images in set $B$. The results shows that, by integrating RC into the objective function, we are able to obtain more accurate CDR values and higher correlations. 
 It is also observed that  SC+RC, SDC+RC and SSGL+RC converges with different speed. SSGL+RC converges fastest after 4 to 5 iterations. SDC+RC requires around 8 iterations and SC+RC requires more iterations to converge.  
 
 In Fig. \ref{fig44}, we show   sample results from two images with different CDRs as well as the reconstruction coefficients by SDC and SDC+RC methods. In Fig. \ref{fig45}, we show the results by SSGL and SSGL+RC methods.  As we can see, by including the RC term into SDC and SSGL, we reduce the CDR range of discs with non-zero reconstruction coefficients and obtain more accurate results.    A limitation of SRCL is that the computational cost increases linearly depending on the number of iterations. Based on our MATLAB implementations in a
 dual core 2.1 GHz PC with 128 GB RAM, it takes 0.27 and
 0.28 second to compute an initial grade using SC and SDC
 method respectively. Each iteration of computing $w_t$ and $\hat{g}_t$
 takes 0.13 second on average for both SC+RC and SDC+RC
 methods. For SSGL+RC, it takes 0.16 second to compute
 an initial grade and each iteration requires 0.035 second on
 average using the same PC.
 
  Table \ref{table1} compares the results with different methods. For simplicity, we use $t=5, 8, 10$ for SSGL+RC, SDC+RC, SC+RC, respectively.  We have also conducted a comparison  with the state-of-the-art segmentation based approaches. Three methods are compared, including the superpixel classification \cite{tmi2013}, the U-Net deep learning \cite{Ronneberger2015} and the latest M-Net deep learning \cite{huazhu18}. To have a fair comparsion, the same sets of
  training and testing images are used. In addition, the original
  implementations are used to get the results. From the results, we can see that the SRCL based method outperforms the segmentation based methods.

Noted that there are several paramters $\lambda_1$, $\lambda_2$, $\lambda_3$ and $\gamma$ in
these methods. A cross-validation within the reference set
has been conducted to determine  $\lambda_1$, $\lambda_2$, $\lambda_3$ in \cite{CJ15}, \cite{Cheng:17BOE}. We
use the same settings: For SC, SC+RC, SDC and SDC+RC, $\lambda_1$ does not need a manually determined value. As discussed
in the paper, they are essentially the same the standard $\ell_1$-norm regularized least square minimization problem, which is
solved using LARS \cite{lars2004}. In LARS, this problem is converted
into the following constrained optimization problem:
 \begin{equation}
\arg \min_{w} (\|\hat{\textbf{y}}-\hat{X}\textbf{w} \|^2), s.t., \|\textbf{w} \|_1 < s,
	\end{equation}
  where $s$ is inversely related to $\lambda_1$. Both $s$ and $\textbf{w}$ are updated 
 iteratively in the LARS. Previous experience \cite{CJ15} shows that
 an iteration of 100 time (i.e, $s=100$) in LARS is able to obtain
a good result and this paper uses 100 iterations for all the $\ell_1$-norm regularized least square minimization problem.
For SDC and SDC+RC, $\lambda_2$ is set as $10^4$.
In SSGL and SSGL+RC, we
 set $\lambda_1$=0.01, $\lambda_2$=10 and $\lambda_3$=0.05.
$\gamma$ is a new parameter to
control the weight of the RC term. In SC+RC and SDC+RC,
$\gamma$=200. In SSGL+RC, 
$\gamma$=100. Our experience shows that a
small change of the paramerters does not affect much the
results.
\begin{table}
  \caption{ Performance comparison of SRCL for   CDR computation.
} \begin{center}
        \begin{tabular}{c|c|c   } \hline
     & {CDR Error}  &    Correlation        
               \\\hline
         	Superpixel \cite{tmi2013} & 0.0774 & 0.590 \\ \hline
         U-net \cite{Ronneberger2015}  & 0.0740  & 0.532  \\ \hline
         M-net \cite{huazhu18} & 0.0710  &  0.586 \\ \hline
             SC \cite{sparsecoding}   & 0.0696& 0.667       \\\hline
           
                       SDC \cite{CJ15}  & 0.0660  &    0.686         \\\hline
                     SSGL \cite{Cheng:17BOE} & 0.0616 & 0.700 \\\hline
                        SC+RC & 0.0669  & 0.671    \\ \hline
                           SDC + RC  &  0.0613    &    0.704        \\\hline
                SSGL +RC & \textbf{0.0599}  & \textbf{0.708}  \\\hline
        \end{tabular}
    \end{center}
 \label{table1}
\end{table}

\subsection{Cataract grading}

\subsubsection{Dataset} We use the ACHIKO-NC dataset \cite{5415679}. It has 5378 images with both decimal  grading score  based on Wisconsin protocol \cite{Klein1990}. By taking the ceiling of the decimal scores,   ACHIKO-NC has 94 images of integral score 1, 1874 images of  integral score 2, 2476 images of  integral score 3, 897 images of  integral score 4 and 37 images of integral score 5. In the experiments, we use the decimal score as manual ground truth in the calculation. 
\subsubsection{Evaluation Metrics}
Following the previous work  \cite{5415679}, we use   four evaluation criteria   to measure the grading accuracy:
\begin{itemize} 
	\item $\epsilon$: the mean absolute error\begin{equation}
	\epsilon=\frac{1}{N} \sum |{G_i}-{M_i}|
	\end{equation}
	\item   $R_0$: the integral agreement ratio \begin{equation}
	R_0= \frac{1}{N} \big{\|}\lceil{G_i}\rceil=\lceil{M_i}\rceil \big{\|}_0
	\end{equation}
	\item  $R_{0.5}$: the percentage of decimal grading with error $\leq 0.5$ \begin{equation}
	R_{0.5}= \frac{1}{N} \big{\|}|\lceil{G_i}\rceil-\lceil{M_i}\rceil|\leq 0.5\big{\|}_0
	\end{equation}
	\item $R_{1}$: the percentage of decimal grading with error  $\leq 1$  \begin{equation}
	R_{1}= \frac{1}{N} \big{\|}|\lceil{G_i}\rceil-\lceil{M_i}\rceil|\leq 1\big{\|}_0
	\end{equation}
	
\end{itemize}
Here $G_i$  denotes the automatically computed grade for the $i^{th}$ image, and $M_i$  denotes manual ground truth grade, $\lceil \cdot \rceil$ denotes the ceiling function,   $\|\cdot\|_0$ denotes the $\ell_0$ norm. 
 
 In cataract grading,  the lens is first detected by the active shape model \cite{5415679} and divided into three parts including the nucleus part, anterior cortex part and posterior cortex part.  The  bag-of-words (BoW) model \cite{bow} is used to extract features   from the three different parts. 
 The BoW model provides a location-independent global representation of local features  
 which are invariant to    rotation, scaling or affine transform.  In cataract grading, the local features in our BoW model are image patches that represent  intensity and texture information. Each part of the resized lens image is divided into  a grid of half-overlapping $s\times s$  patches and $s=3$ in our experiments.   
 After obtaining all the local patches from a set of training images, $k$-means clustering technique is applied to generate the codebook from randomly selected samples, and then the BoW is computed  from a binning procedure, where the clustering parameter $k$ is referred to as the bin number.
 More details of BoW computation can be found in \cite{bow}.
  
\begin{figure}
	\centering
	
	\includegraphics[width=3.0in]{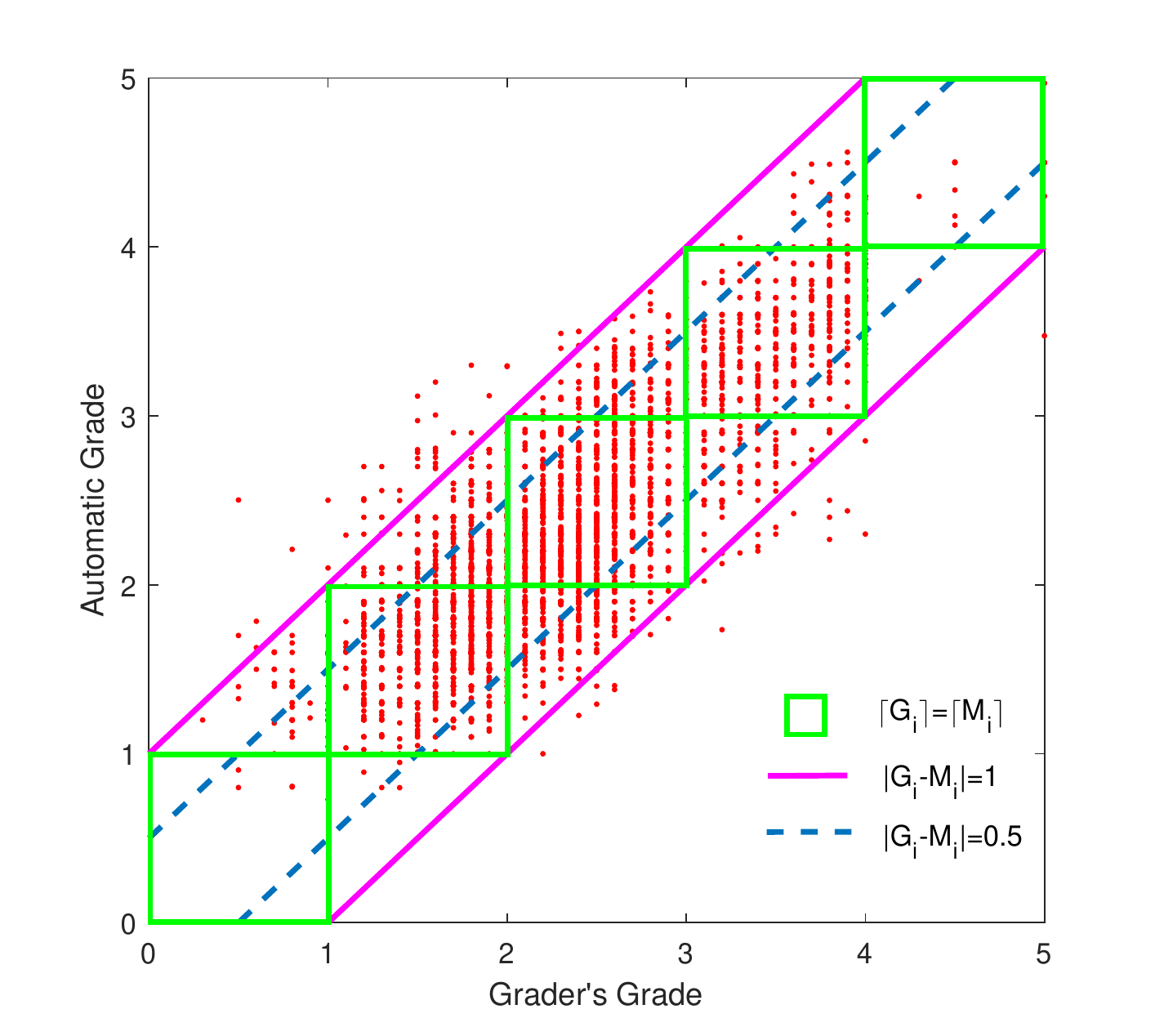}

	\caption{Visual illustration of the agreement betweeen automatic grading by SSGL+RC and the Grader's grading.} \label{fig67}
\end{figure}

We use a randomly selected 340 images  as reference data,  where the number of images from grade 1 to 5 are 20, 100, 100, 100, 20 respectively. We use fewer images from grade 1 and 5 because the number of images in these two grades are much smaller than other grades. 
  The remaining 5038 images are used for testing. Similar to that in CDR computation, we integrate RC with SC, SDC and SSGL to get the results by SC+RC, SDC+RC and SSGL+RC. A cross-validation within the reference set is conducted to determine the parameters. 
  Similarly, $\lambda_1$ does not need to be determined in LARS for
  SC, SDC, SC+RC, SDC+RC. For SDC and SDC+RC, we set $\lambda_2$=2. For SSGL and SSGL+RC, 
  $\lambda_1$=0.035,$\lambda_2$=10,$\lambda_3$=0.05. The new parameter $\gamma$=100 in all the experiments.  
    The results are summarized in Table \ref{table3}. In addition, we have also compared the proposed method  with other methods. Two algorithms are compared, including the RBF  kernel-based support vector regression (RBF+SVR) approach \cite{5415679}   and    group lasso regression approach (GSR) \cite{Xu13}.   As we can see, the proposed method outperforms  other methods.
 In Fig. \ref{fig67}, we also plot the automatic grading by SSGL+RC against that of a professional grader,  which demonstrates that the proposed method has good agreement with the professional grader. 
\begin{table}
	\caption{ Performance comparison of SRCL for cataract grading.
	} \begin{center}
		\begin{tabular}{c|c|c  |c|c } \hline
			& {$\epsilon$}  &   $R_0$ & $R_{0.5}$ & $R_1$
			\\\hline
			RBF+SVR \cite{5415679}  & 0.363 & 0.615 & 0.711 & 0.980 \\ \hline
			BoW+GSR \cite{Xu13} & 0.356 & 0.679 & 0.750   & 0.983 \\ \hline
	    	LLC \cite{yanwu2016} & 0.331 & 0.688 & 0.767 & 0.987 \\ \hline
			SC \cite{sparsecoding}   & 0.367& 0.666 & 0.724 & 0.980       \\\hline
			
			SDC \cite{CJ15}  & 0.350  &    0.681 & 0.741 & 0.986         \\\hline
				SSGL \cite{Cheng:17BOE}    & 0.339  &    0.677 & 0.758 & 0.986         \\\hline
			SC+RC & 0.329 & 0.687 & 0.761 & 0.987    \\\hline
			SDC + RC  & {0.325}   &   {0.691} &  {0.771} &  {0.988}       \\\hline
				SSGL+RC \   &  \textbf{0.322}  &     \textbf{0.691} &  \textbf{0.780} &  \textbf{0.989}        \\\hline
			
		\end{tabular}
	\end{center}
	\label{table3}
\end{table}
%

\section{Discussions and conclusions} \label{discuss}
In this paper, we propose a new regularization term named range constraint and combine it with sparse learning for SRCL. The proposed SRCL is able to improve the automatic grading of CDR from retinal fundus images and cataract severity from slit-lamp lens images.  Here we discuss a little more why it works and how we can extend it for other applications. Recall that in the grading, we want to find some atoms from the dictionary to reconstruct a testing image $\textbf{y}$. However, there is no perfect data representation  (either the original image representation or a feature map computed from the image), such that a distance $d$ computed from the data perfectly reflects the grading of the image. Therefore,  some atoms may have high weights in $\textbf{w}$, though they are not really similar to $\textbf{y}$ in the grading. By introducing the range-constraint, we avoid to reconstruct the testing images with atoms that have quite different grades.

The method can be extended for other medical image grading
applications and generic image classification or grading
problems. In general, it is suitable for a grading problem which
aims to find a decimal score for a new image based on a set
of atoms with known scores. The data representation can be
from pixel intensities of the images or feature maps computed
from the images. But the intensities or the feature maps shall
have the property that those images similar in the grading
scores shall be similar in the data representation. Therefore,
we need to find appropriate data representation when applying
the model to other grading problems for medical or generic
images. A potential application is the age estimation from
facial images. To make the algorithm applicable for age
estimation, one needs to compute a feature representation of
the images such that the features of images from similar ages
are often close. We will explore the use of the algorithm for
such application in future work. It should be noted that for
a grading problem without decimal scores as ground truth,
SRCL may not have much advantages as the rounding error
may be introduced to the grading.

In this paper, we propose a novel sparse range-constrained
learning method for medical image grading. It combines the
objective of sparse learning with the objective of medical
image grading into one function. Our results show that such
a combination is effective and leads to better performance
in both CDR computation and cataract grading. Our method
is a general approach and it can be extended for automatic
grading of other diseases. A limitation of our approach is that
it requires more computational cost. A second limitation is
that the method is not appliable for a problem that cannot be
solved well by a sparse coding based method. Noise will also
affect the performance of the method. In case of applications
where the images have moderate or high level of noise, we
may need to develop or apply some algorithms to remove the
noise before or during the feature extraction.

\bibliographystyle{IEEEbib}
\bibliography{bibdata}

\end{document}